\documentclass[conference]{IEEEtran}
\IEEEoverridecommandlockouts
\makeatletter
\def\ps@IEEEtitlepagestyle{%
  \def\@oddfoot{%
    \parbox{\textwidth}{\scriptsize
  
    Copyright © 2025 IEEE. All rights reserved. This is the accepted version of the paper. The final version will be published in the \textit{IEEE International Conference on Intelligent Transportation Systems (ITSC), 2025}}
    }%
  }%

\makeatother

\usepackage{url}
\usepackage{svg}
\usepackage{cite}
\usepackage{amsmath,amssymb,amsfonts}
\usepackage{algorithm}
\usepackage{graphicx}
\usepackage{textcomp}
\usepackage{xcolor}
\usepackage{booktabs}
\usepackage{svg}
\usepackage{algpseudocode}
\usepackage{multirow}
\usepackage{hyperref}
\usepackage{makecell}
\newtheorem{thm}{Problem}

\def\BibTeX{{\rm B\kern-.05em{\sc i\kern-.025em b}\kern-.08em
    T\kern-.1667em\lower.7ex\hbox{E}\kern-.125emX}}

\usepackage{amsmath,amsfonts,bm}
\usepackage{xspace}

\newcommand{\bmu}{{\boldsymbol{\mu}}}

\newcommand{\model}{Path-Diffuser}

\def\eqref#1{equation~\ref{#1}}
\def\Eqref#1{Equation~\ref{#1}}

\def\1{\bm{1}}

\def\rvd{{\mathbf{d}}}

\def\rvh{{\mathbf{h}}}

\def\rvl{{\mathbf{l}}}

\def\rvs{{\mathbf{s}}}

\def\rvx{{\mathbf{x}}}

\def\rmI{{\mathbf{I}}}

\DeclareMathAlphabet{\mathsfit}{\encodingdefault}{\sfdefault}{m}{sl}
\SetMathAlphabet{\mathsfit}{bold}{\encodingdefault}{\sfdefault}{bx}{n}

\def\gC{{\mathcal{C}}}

\def\gL{{\mathcal{L}}}

\def\gN{{\mathcal{N}}}

\def\gU{{\mathcal{U}}}

\def\sD{{\mathbb{D}}}

\def\sV{{\mathbb{V}}}

\newcommand{\fig}[1]{Fig.~\ref{#1}}

\usepackage[acronym]{glossaries}
\newacronym{SGD}{\textsc{sgd}}{stochastic gradient descent}
\newacronym{MAP}{\textsc{map}}{maximum-a-posteriori}
\newacronym{MLE}{\textsc{mle}}{maximum likelihood estimation}
\newacronym{MNLL}{\textsc{mnll}}{mean negative log-likelihood}
\newacronym{NLL}{\textsc{nll}}{negative log-likelihood}
\newacronym{LL}{\textsc{ll}}{log-likelihood}
\newacronym{RMSE}{\textsc{rmse}}{root mean square error}
\newacronym{ECE}{\textsc{ece}}{expected calibration error}
\newacronym{SNR}{\textsc{snr}}{signal-to-noise ratio}
\newacronym{FID}{\textsc{fid}}{Fr\'echet Inception Distance}
\newacronym{BPD}{\textsc{bpd}}{bit per dimension}
\newacronym{NFE}{\textsc{nfe}}{neural function evaluations}

\newacronym{AE}{\textsc{ae}}{auto-encoder}
\newacronym{WAE}{\textsc{wae}}{Wasserstein Auto-encoder}
\newacronym{VAE}{\textsc{vae}}{Variational Auto-encoder}
\newacronym{BAE}{\textsc{bae}}{Bayesian Auto-encoder}
\newacronym{CDF}{\textsc{cdf}}{cumulative density function}
\newacronym{GAN}{\textsc{gan}}{Generative Adversarial Network}
\newacronym{DPGMM}{\textsc{dpgmm}}{Dirichlet process Gaussian mixture model}
\newacronym{GMM}{\textsc{gmm}}{Gaussian mixture model}

\newacronym{MC}{mc}{Monte Carlo}
\newacronym{SDE}{\textsc{sde}}{Stochastic Differential Equation}
\newacronym{CNF}{cnf}{Continuous Normaxlizing Flow}
\newacronym{ODE}{ode}{Ordinary Differential Equation}
\newacronym{MCMC}{\textsc{mcmc}}{Markov chain Monte Carlo}
\newacronym{HMC}{\textsc{hmc}}{Hamiltonian Monte Carlo}
\newacronym{MH}{mh}{Metropolis-Hastings}
\newacronym{NUTS}{nuts}{no-u-turn sampler}
\newacronym{SGHMC}{\textsc{sghmc}}{stochastic gradient Hamiltonian Monte Carlo}

\newacronym[longplural=deep Gaussian processes]{DGP}{\textsc{dgp}}{deep Gaussian process} 
\newacronym{GPLVM}{gplvm}{Gaussian process latent variable model}
\newacronym{DPMM}{dpmm}{Dirichlet Process Mixture Model}

\newacronym{VFE}{vfe}{variational free energy}

\newacronym[longplural=Gaussian Processes]{GP}{\textsc{gp}}{Gaussian Process}

\newacronym{VI}{\textsc{vi}}{variational inference}
\newacronym{SVI}{\textsc{svi}}{stochastic variational inference}

\newacronym{ELBO}{\textsc{elbo}}{evidence lower bound}
\newacronym{NELBO}{\textsc{nelbo}}{negative evidence lower bound}
\newacronym{ELL}{\textsc{ell}}{expected log likelihood}
\newacronym{KL}{\textsc{kl}}{Kullback-Leibler}
\newacronym{AUC}{auc}{area under the curve}

\newacronym{BNN}{\textsc{bnn}}{Bayesian neural network}
\newacronym{DNN}{\textsc{dnn}}{deep neural network}
\newacronym{CNN}{\textsc{cnn}}{convolutional neural network}
\newacronym{MLP}{\textsc{mlp}}{multilayer perceptron}
\newacronym{NN}{nn}{neural network}
\newacronym{RELU}{ReLU}{rectified linear unit}

\newacronym{NF}{nf}{normalizing flow}

\newacronym{RBF}{rbf}{radial basis function}
\newacronym{ARD}{ard}{automatic relevance determination}

\newacronym{RKHS}{rkhs}{reproducing kernel Hilbert space}

\newacronym{OT}{ot}{optimal transport}
\newacronym{WD}{wd}{Wasserstein distance}
\newacronym{SWD}{swd}{sliced-Wasserstein distance}
\newacronym{DSWD}{dswd}{distributional sliced-Wasserstein distance}
\newacronym{fsp}{FSP}{Fictitious Self Play}
\newacronym{marl}{MARL}{Multi-agent reinforcement learning}
\newacronym{pomg}{POMG}{Partially Observable Markov Games}
\newacronym{ddpm}{DDPM}{Denoising Diffusion Probabilistic Models}
\newacronym{rl}{RL}{Reinforcement Learning}
\newacronym{sl}{SL}{Supervised Learning}
\newacronym{dp}{DP}{Diffusion Policy}

\newacronym{mpe}{MPE}{Multiple Particle Environment}
\newacronym{pd}{PD}{{Path Diffuser}}
\newacronym{pd_p}{PD$\ominus$P}{PD without Primitives}
\newacronym{dt}{DIFFT}{Differential Transformer}
\newacronym{ff}{Frenet Frame}{The Frenet-Serret Frame}
\newacronym{mha}{MHA}{Multi-Head Attention}
\newacronym{hmp}{HMP}{Heterogeneous Message Passing}
\newacronym{vd}{VD}{Vanilla Diffusion}
\newacronym{cdb}{CDB}{Centralized and Decentralized Behavior}

\begin{document}

\title{
Path Diffuser: Diffusion Model for Data-Driven Traffic Simulator}

\author{
    Da Saem Lee, Akash Karthikeyan, Yash Vardhan Pant, Sebastian Fischmeister
    
    \thanks{This work is supported in part by the Natural Sciences and Engineering
Research Council of Canada (NSERC) and Canada Foundation for Innova-
tion - John R. Evans Leaders Fund (CFI JELF).}
    \thanks{The authors are with the Department of Electrical and Computer Engineering, University of Waterloo, Waterloo, Canada.
  \ttfamily{ds3lee@uwaterloo.ca, a9karthi@uwaterloo.ca, yash.pant@uwaterloo.ca, sfischme@uwaterloo.ca}
}
}

\maketitle

\begin{abstract}
  
Simulating diverse and realistic traffic scenarios is critical for developing and testing autonomous planning. Traditional rule-based planners lack diversity and realism, while learning-based simulators often replay, forecast, or edit scenarios using historical agent trajectories. However, they struggle to generate new scenarios, limiting scalability and diversity due to their reliance on fully annotated logs and historical data.
%
Thus, a key challenge for a learning-based simulator's performance is that it requires agents' past trajectories and pose information in addition to map data, which might not be available for all agents on the road. Without which, generated scenarios often produce unrealistic trajectories that deviate from drivable areas, particularly under out-of-distribution (OOD) map scenes (e.g., curved roads). 
To address this, we propose \gls{pd}: a two-stage, diffusion model for generating agent pose initializations and their corresponding trajectories conditioned on the map, free of any historical context of agents’ trajectories.  Furthermore, \gls{pd} incorporates a motion primitive-based prior, leveraging Frenet frame candidate trajectories to enhance diversity while ensuring road-compliant trajectory generation. We also explore various design choices for modeling complex multi-agent interactions. 
We demonstrate the effectiveness of our method through extensive experiments on the Argoverse2 Dataset and additionally evaluate the generalizability of the approach on OOD map variants. Notably, \model{} outperforms the baseline methods by $1.92\times$ on distribution metrics, $1.14\times$ on common-sense metrics, and $1.62\times$ on road compliance from adversarial benchmarks\footnotemark
\footnotetext{\href{https://github.com/CL2-UWaterloo/PathDiffuser/}{\textcolor{blue}{https://github.com/CL2-UWaterloo/PathDiffuser/}}}.

\end{abstract}

\section{Introduction}
Traffic scenario simulations that reflect the real world are crucial for evaluating autonomous driving systems \cite{fremont2020formal}. Scenario simulation can be decomposed into two steps: (1) agent pose initialization and (2) trajectory generation. While trajectories can technically originate from any location on the map, realistic agent initialization is necessary to reflect real-world behavior. For instance, at the intersection in the real world, some vehicles may stop behind the stop line to allow other vehicles to pass through. Further, given an initialization, the trajectories should be generated by following the traffic rules (e.g., driving on the correct side of the road) and not colliding with the other agents to follow a real-world scenario.

Industry practitioners manually specify scenarios for scenario-based testing \cite{antkiewicz2020modes, fremont2020formal} and use vehicle simulators, such as SUMO~\cite{SUMO2018} and CARLA~\cite{dosovitskiy2017carlaopenurbandriving}. Although these frameworks are helpful, they often lack the realism and diversity necessary to fully reflect real-world conditions. Recent works \cite{lu2024scenecontrol, jiang2023motiondiffusercontrollablemultiagentmotion, wang2024optimizingdiffusionmodelsjoint, rowe2025scenariodreamer} have adopted learning-based techniques to mitigate this limitation and capture the underlying distributions in complex and diverse traffic scenarios.

Although data-driven traffic simulators have made significant progress with rasterized representation of a scene~\cite{pronovost2023scenario, Chitta2024ECCV, sun2024drivescenegen}, several challenges remain, including scalability to the longer horizon and the complexity to incorporating control signals or constraints. To address these limitations, recent works~\cite{ jiang2023motiondiffusercontrollablemultiagentmotion, wang2024optimizingdiffusionmodelsjoint, rowe2025scenariodreamer} utilizes vectorized representation of a scene, which offer a more compact, interpretable, and flexible encoding of map and agent features.
However, these methods often depend on agent history to generate future trajectories, leading to models that overfit and replay logged behaviors rather than generalizing to diverse scenarios. Moreover, absense of historical trajectories of other road agents in driving logs makes it challenging to scale these approaches.
In the absence of historical context, generated scenarios often produce unrealistic behaviors, such as the vehicles stay still or deviate from drivable areas, particularly under out-of-distribution (OOD) map conditions (e.g., curved roads). 

Agent initialization remains a key challenge. A traffic scenario can be generated by randomly placing vehicles in the scene, though this raises concerns about realism. In order to generate realistic traffic scenario, it is critical to have realistic initialization that aligns with the map structure and meets common sense, such as agents inside the drivable area or not in collision. To address this, recent work~\cite{lu2024scenecontrol} utilizes the transformer to generate agent initializations given a map. However, this method requires the guidance sampling for collision and map constraints to generate realistic placements.

\noindent\textbf{Contributions of this work.} To address these challenges, we propose \model{}, a diffusion-based framework for traffic scenario generation that models agent initialization and trajectory generation. Our key contributions are as follows:

\begin{itemize}
    \item We introduce a two stage diffusion framework that jointly generates agent initializations and their trajectories, conditioned on the map without requiring historical data.

   \item For Agent Initialization, we incorporate the differential transformer to suppress attention noise and generate realistic, spatially consistent scene configurations by attending relevant map context.
    \item For trajectory generation, we introduce Frenet-frame candidates, enabling the model to capture diverse trajectories that are more robust to map perturbations.
\end{itemize}
In this framework, trajectory generation is designed to be independent of historical context, as such information may not be available for every agent in the scene.
Moreover, we validate our framework on the Argoverse v2 dataset through qualitative and quantitative evaluation. To demonstrate the robustness to the OOD map variants, we also evaluated on the perturbed map. 

\noindent \textbf{Outline of the paper.} Section~\ref{sec:relatedworks} provides a brief overview of existing methods for agent initialization, motion forecasting, and end-to-end approaches for traffic scenario generation and simulation. Section~\ref{sec:preliminaries} presents the required background, notation, and introduces the problem statement for this work. Section~\ref{sec:approach} describes the proposed agent initialization and trajectory generation approaches. Section~\ref{sec:experiments} presents quantitative and qualitative evaluations that benchmark the performance of the proposed approach against baselines. Finally, we discuss limitations and future work in Section~\ref{sec:conclusion}.

\section{Related Works}
\label{sec:relatedworks}

Although rule-based traffic simulation methods, such as \cite{antkiewicz2020modes, fremont2020formal}, are widely used for testing autonomous driving systems, given the scope of this work, we restrict this section to data-driven methods for traffic scenario generation and simulation. 

\paragraph{Initialization of a traffic scenario}  
SceneGen~\cite{tan2021scenegen} and SceneControl~\cite{lu2024scenecontrol} generate agent bounding boxes, while other approaches~\cite{Chitta2024ECCV, sun2024drivescenegen} generate both agent boxes and lane graphs. However, SLEDGE~\cite{Chitta2024ECCV} and DriveSceneGen~\cite{sun2024drivescenegen} rely on rasterized images for initialization, resulting in parameter-heavy architectures. In contrast, SceneControl proposes a controllable scene generation framework that produces initial 2D poses and vehicle attributes (e.g., size, speed) using guided diffusion sampling~\cite{dhariwal2021diffusion}. SceneControl enforces realism and controllability through constraints such as collision avoidance and off-lane penalties. However, without careful balancing of these constraints, performance degrades. In contrast, we aim to generate initial scenes to produce more realistic scenarios without relying on guided sampling.

\paragraph{Motion Forecasting} For autonomous vehicles, predicting the motion of non-ego agents is crucial for safe navigation. Consequently, extensive research has focused on trajectory prediction. Similar to next token prediction for language model, Trajeglish~\cite{philion2024trajeglishtrafficmodelingnexttoken} formulates trajectory forecasting as a token prediction task using a trajectory codebook. Similarly, MotionDiffuser~\cite{jiang2023motiondiffusercontrollablemultiagentmotion} shows that representing trajectories in a compact and expressive latent space enables efficient and high-quality trajectory generation. Building on MotionDiffuser, OptTrajDiff~\cite{wang2024optimizingdiffusionmodelsjoint} adapts latent diffusion and introduces Optimal Gaussian Diffusion, which incorporates marginal mode predictions from QCNet~\cite{zhou2023query} as a diffusion prior to reduce the number of diffusion steps. By leveraging QCNet predictions, OptTrajDiff achieves approximately $5\times$ computational efficiency compared to other diffusion-based baselines. However, these approaches rely on past trajectories, which may not be available for all road agents. In their absence, generated scenarios often exhibit unrealistic behaviors, or fail to produce novel and diverse interactions. In contrast, our method does not depend on historical trajectories and instead generates agent behaviors solely from their initial states and map context.

\paragraph{End-to-End Model for Traffic Scenario Generation} For end-to-end traffic simulation, TrafficGen~\cite{feng2023trafficgen} uses a Gaussian Mixture Model (GMM) to generate initial placements by sequentially placing agents one by one. Trajectory generation is then treated as a motion-forecasting task, conditioned on the agents' historical context. SceneDiffuser~\cite{jiang2024scenediffuserefficientcontrollabledriving} adopts a diffusion-based model to generate full traffic scenes end-to-end, supporting operations such as perturbing scenes while preserving realism, injecting agents, and synthesizing new scenarios with realistic layouts. Additionally, SceneDiffuser introduces a protocol for scalable scenario generation by incorporating constraint specifications derived from Large Language Models (LLMs). However, to generate trajectories for newly initialized agents, it still requires partial scene information, such as the state or history of other agents.

\section{Problem Statements and Preliminaries}
 \label{sec:preliminaries}

\subsection{Notations}
We begin by defining the initial state of each traffic agent $i$ as a vector $\vec{x}_i = [x_i, y_i, \theta_i, v_i, c_i] \in \mathbb{R}^5$, where $(x_i, y_i) \in \mathbb{R}^2$ denotes the initial 2D position, $\theta_i \in [-\pi, \pi)$ the heading angle, $v_i \in \mathbb{R}_+$ the velocity, and $c_i \in \mathbb{Z}_+$ the agent type. The initial states of all $N$ agents are denoted as $\vec{\rvx} = \{\vec{x}_1, \ldots, \vec{x}_N\} \in \mathbb{R}^{N \times 5}$. Each agent’s trajectory over horizon $H$ is $\tau_i = \{(x,y)_{i,1}, \ldots, (x,y)_{i,H}\}$, and the joint trajectory is $\boldsymbol{\tau} = \{\tau_1, \ldots, \tau_N\} \in \mathbb{R}^{N \times H \times 2}$. Let $\mathcal{M}$ denote the map, a set of 2D map points $m \in \mathbb{R}^2$, and lanes $l \in \mathbb{R}^{K \times 2}$, each represented as an ordered sequence of $K$ such map points.   
Diffusion steps $t$ are denoted by superscripts, e.g., $\vec{x}_i^t$ with $t \in \{1, 2, \dotsc, T\}$. We use $\mathcal{U}[1,T]$ for the uniform distribution over $\{1, \dotsc, T\}$, and $\mathcal{N}$ for the Gaussian distribution. The notation $\sim$ indicates sampling, e.g., $\epsilon \sim \mathcal{N}(0, I)$.

\subsection{Problem Statement}

Given a map $\mathcal{M}$ and the number of agents in the map, we aim to develop a generative model that produces traffic scenarios aligned with the real-world distributions $q(\Vec{\rvx}^0)$ and $q(\boldsymbol{\tau}^0)$, corresponding to agent initializations and their subsequent trajectories, respectively. To this end, we decompose the problem into two sub-tasks: 

\begin{thm}\label{prob:init}
(Scene Initialization) 
Given a map $\mathcal{M}$ and number of agents $N$, the goal is to learn a generative model that samples initial states $\hat{\Vec{\rvx}} \sim \Vec{\rvx}^0$, such that the generated samples resemble the real data.
\end{thm}

\begin{thm}\label{prob:traj}
(Trajectory Generation) 
Given a map $\mathcal{M}$ and agent initializations ${\Vec{\rvx}^0}$, we aim to learn a generative model that produces trajectories $\boldsymbol{\tau}$ consistent with the underlying road structure and behaviors.
\end{thm}

\begin{figure*}[t]
    \centering
    \includegraphics[width=0.9\linewidth]{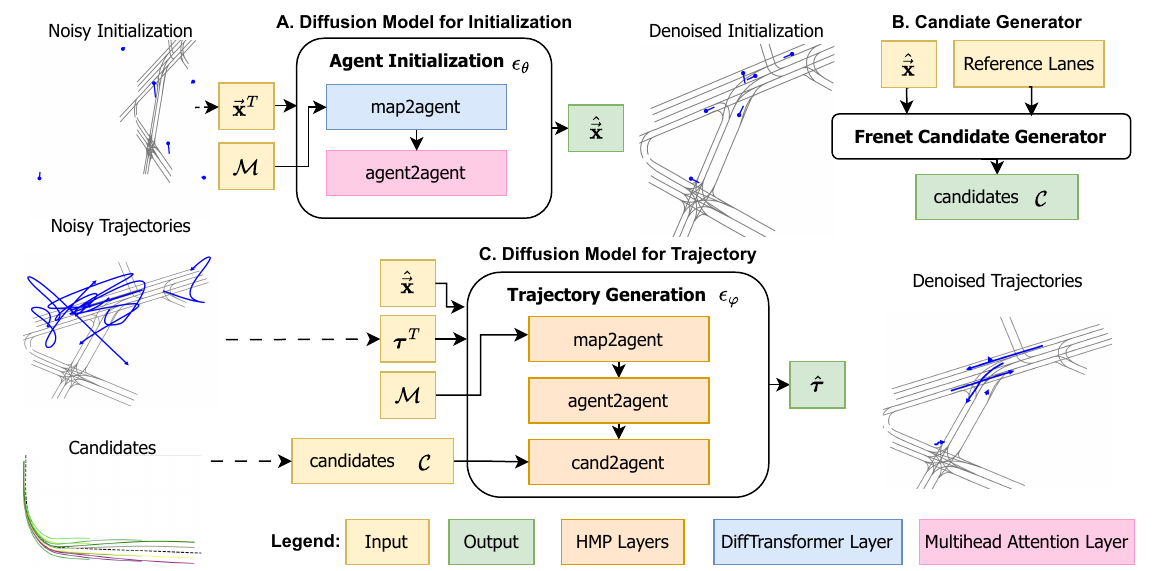}
    \caption{\textbf{Overview of \model}. In \textbf{(A)}, Agent initialization takes noisy initialization $\Vec{\rvx}^T$ and predicts the noise $\bm\epsilon_\theta$ (See Algorithm \ref{alg:init_training}). Frenet candidate trajectories are generated in \textbf{(B)} based on the initial pose as a prior (details in Algorithm \ref{alg:frenet_cand}). \textbf{(C)} Trajectory denoiser takes the initialization and noisy latent trajectory $\bm\tau^T$ and denoises by taking the candidate trajectories.} 
    \label{fig:overall_diagram}
\end{figure*}
\subsection{Diffusion Model}
Denoising diffusion probabilistic models (DDPM)~\cite{ho2020denoisingdiffusionprobabilisticmodels} are an expressive class of generative models. Latent diffusion models~\cite{rombach2021highresolution} extend this framework by learning a generative model in latent space, reducing computational overhead. The diffusion framework transforms the data distribution into an isotropic Gaussian (prior) through a forward process, then iteratively denoises the samples from prior distribution to recover the original data through the reverse process.

\noindent\textbf{Forward Process.} Let $q(\Vec{\rvx}^0)$ and $q(\boldsymbol{\tau}^0)$ denote the data distributions of agent initial states and trajectories, respectively. Gaussian noise is incrementally added over $T$ steps according to a variance schedule $\beta_1, \dotsc, \beta_T$, such that the samples $\Vec{\rvx}^T, \boldsymbol{\tau}^T$ approach $\mathcal{N}(\mathbf{0}, \mathbf{I})$. The transition kernels at each step are defined as:
\begin{align}
 \label{eq:init_forward}
q(\Vec{\rvx}^t | \Vec{\rvx}^{t-1}) &:= \mathcal{N}\left(\Vec{\rvx}^t; \sqrt{1 - \beta_t} \, \Vec{\rvx}^{t-1}, \, \beta_t \, \mathbf{I} \right), \\
q(\boldsymbol{\tau}^t | \boldsymbol{\tau}^{t-1}) &:= \mathcal{N}\left(\boldsymbol{\tau}^t; \sqrt{1 - \beta_t} \, \boldsymbol{\tau}^{t-1}, \, \beta_t \, \mathbf{I} \right),
\end{align}

\noindent
By the Markov property of the forward process, the marginal distribution of the noisy samples at any step $t$ can be expressed as a conditional distribution given the original data:
\begin{align}
\label{eq:reverse}
q(\Vec{\rvx}^t | \Vec{\rvx}^0) &= \mathcal{N}\left(\Vec{\rvx}^t; \sqrt{\bar{\alpha}_t} \, \Vec{\rvx}^0, (1 - \bar{\alpha}_t) \, \mathbf{I} \right), \\
q(\boldsymbol{\tau}^t | \boldsymbol{\tau}^0) &= \mathcal{N}\left(\boldsymbol{\tau}^t; \sqrt{\bar{\alpha}_t} \, \boldsymbol{\tau}^0, (1 - \bar{\alpha}_t) \, \mathbf{I} \right),
\end{align}

\noindent
where $\alpha_t = 1 - \beta_t$ and $\bar{\alpha}_t = \prod_{s=1}^{t} \alpha_s$. Using the reparameterization trick, we can sample noisy input at any $t$ as follows:
\begin{equation}
\label{eqn:reparam}
\Vec{\rvx}^t = \sqrt{\bar{\alpha}_t} \, \Vec{\rvx}^0 + \sqrt{1 - \bar{\alpha}_t} \, \boldsymbol{\epsilon}, \quad
\boldsymbol{\tau}^t = \sqrt{\bar{\alpha}_t} \, \boldsymbol{\tau}^0 + \sqrt{1 - \bar{\alpha}_t} \, \boldsymbol{\epsilon},
\end{equation}

\noindent

\noindent\textbf{Reverse Process.} 
The joint distribution $p_\theta(\Vec{\rvx}_{0:T})$ defines the reverse process. Starting from the prior distribution $p_\theta(\Vec{\rvx}^T) = \mathcal{N}(\Vec{\rvx}^T; \mathbf{0}, \mathbf{I})$, the model learns to denoise by parameterizing the transitions as  
$p_\theta(\Vec{\rvx}^{t-1}| \Vec{\rvx}^t) = \mathcal{N}(\Vec{\rvx}^{t-1}; \bmu_\theta(\Vec{\rvx}^t, t), \sigma_t^2 \mathbf{I})$, where~${\sigma_k^2 = \beta_t \frac{1-\bar{\alpha}_{t-1}}{1-\bar{\alpha}_t}}$,
to approximate the true posterior $q(\Vec{\rvx}^{t-1} | \Vec{\rvx}^t, \Vec{\rvx}^0)$. Similarly, the reverse process for trajectories can be parameterized as  
$p_\varphi(\boldsymbol{\tau}^{t-1}|\boldsymbol{\tau}^t) = \mathcal{N}(\boldsymbol{\tau}^{t-1}; \bmu_\varphi(\boldsymbol{\tau}^t, t), \sigma_t^2 \mathbf{I})$. Here, $\theta$ and $\varphi$ denote the learnable parameters.

\subsection{Frenet Frame based Candidate Trajectories}
\gls{ff}~\cite{frenet1852courbes} defines a local, path-relative coordinate system in which a vehicle’s position is described by the arc-length $s$, the distance traveled along the reference lane $l_\mathcal{R}\in \mathcal{M}$ and the lateral offset $d$, representing the deviation from the centerline. This representation is invariant under the action of the special Euclidean group $\mathrm{SE}(2) = \mathrm{SO}(2) \times \mathbb{R}^2$.
Naturally adapting to curvy lanes as in~\fig{fig:frenet}, the Frenet Frame simplifies trajectory generation and is widely used in motion planning~\cite{trauth2024frenetix, huang2023trajectory}. 

A trajectory in the Frenet frame is defined by:

\begin{equation}
s(h) = s_0 + v h, \quad
d(h) = d_0 + a_3 h^3 + a_4 h^4 + a_5 h^5
\label{eq:frenet}
\end{equation}
\noindent
where $(s_0, d_0)$ denotes the initial position in the Frenet frame with respect to the reference lane, $h$ is time step, and $a_i$ are the coefficients of the quintic polynomial that govern the lateral deviation. We assume a constant longitudinal velocity $v$, which enables efficient exploration over a grid of initial speeds and lateral offsets to generate a diverse set of feasible trajectories originating from the Cartesian point $(x_i, y_i)$.

\section{\model: Approach}
\label{sec:approach}
\noindent An overview of our method is shown in Fig.~\ref{fig:overall_diagram}. Building on the concepts introduced in Section~\ref{sec:preliminaries}, we now describe our overall framework and its components in detail.
\subsection{Scene Initialization}
\label{sec:scene_init}
\noindent
Given samples from driving logs, our goal is to learn the underlying distribution of agent initializations $\Vec{\rvx}$. To address Problem~\ref{prob:init}, we adopt a DDPM-based formulation.
The reverse diffusion process is intractable, and we therefore train a neural network $\bm{\epsilon}_{\theta}(\Vec{\rvx}^t, \mathcal{M}, t) = \frac{\sqrt{1 - \bar{\alpha}_t}}{\beta_t} \left(\Vec{\rvx}^t - \sqrt{\alpha_t}\,\bmu_{\theta}(\Vec{\rvx}^t, \mathcal{M}, t)\right)$ to approximate it. This training maximizes the Variational Lower Bound (VLO)~\cite{ho2020denoisingdiffusionprobabilisticmodels} 
$
\mathcal{L}_{\mathbb{VLO}} = \mathbb{E}_{q(\Vec{\rvx}^{0:T})} \left[ 
\log \frac{p_\theta(\Vec{\rvx}^{0:T})}{q(\Vec{\rvx}^{1:T} | \Vec{\rvx}^0)}
\right]$. We additionally condition the model on map features to guide generation toward valid initializations. In practice, $\mathcal{L}_{\mathbb{VLO}}$ reduces to the following conditional score matching loss:
\begin{equation}  
\label{eq:init_obj}
\min_{\theta} \mathcal{L}_{\mathbb{I}}(\theta) =
\mathbb{E} \left[
\left\| {\boldsymbol\epsilon} - \bm{\epsilon}_\theta\left( 
\Vec{\rvx}^t, \mathcal{M}, t
\right) \right\|^2
\right]
\end{equation}

\subsubsection{\gls{cdb}} 
\label{subsec:cent}
Modeling multi-agent systems requires capturing inter-agent interactions while allowing for decentralized decision-making. Following~\cite{zhu2025madiffofflinemultiagentlearning}, we regularize learning by alternating between centralized and decentralized modes. In the centralized mode, agents attend to one another via a full attention mask $\mathbf{1}_{n \times n}$ to enable coordinated scene generation. In the decentralized mode, attention is masked using the identity matrix $\mathbf{I}_n$, allowing agents to evolve independently.
This stochastic masking strategy is integrated into our training pipeline and reflected in the training objective (Eq.~\ref{eq:init_obj}) and Line 5 of Algorithm~\ref{alg:init_training}.

\begin{table}[b]
\caption{Attention Score Variance Across Diffusion Steps}
\begin{center}
\begin{tabular}{cc|ccc}
\hline
&& \multicolumn{3}{c}{Diffusion Steps}
\\
                Attention Type & Layer & 1                  & 201                & 401                \\ \hline
\multirow{2}{*}{map-agent}   & MHA             & 1.83e-4          & 1.89e-4          & 1.59e-4          \\
                                &  \textbf{DIFFT}           & \textbf{1.82e-2} & \textbf{1.80e-2} & \textbf{1.76e-2} \\ \hline
\multirow{2}{*}{agent-agent} & \textbf{MHA}            & \textbf{2.74e-3} & \textbf{2.65e-3} & \textbf{2.67e-3} \\
                                & DIFFT           &{1.83e-4}         & {1.89e-4}         & {1.59e-4}         \\ \hline
\end{tabular}
\label{tab:variance}

\end{center}

\end{table}


    
\subsubsection{Design Choices}
To model map-agent and agent-agent interactions, we follow~\cite{zhou2023query, lu2024scenecontrol}, encoding the vectorized map with point-level features such as normalized location, heading, and lane curvature and pairwise connections to capture map geometry.
However, as shown in Table~\ref{tab:variance}, standard \gls{mha}~\cite{vaswani2023attentionneed} struggles to prioritize relevant tokens in scenarios with dense map components which is reflected in low variance across diffusion steps for map-agent attention. This indicates that the agent attends indiscriminately to map tokens regardless of spatial locality. Conversely, agent-agent attention exhibits high variance, suggesting that the model selectively focuses on a subset of agents. With existence of the vast amount of map tokens, attention patterns result in noisy and unstable distributions that degrade quality.

To address this, we incorporate \gls{dt}~\cite{ye2024differential}, a transformer variant that cancels attention noise, promotes sparsity and enhances semantic relevance.
Table~\ref{tab:variance} shows that using \gls{dt} for the map-agent attention significantly increases variance of map-agent attention score, emphasizing local spatial context attention.

\subsubsection{Order Enforcing}
To resolve ambiguity from arbitrary permutations in generated agent initializations, we impose a canonical left-to-right, top-to-bottom ordering and add sinusoidal positional embeddings after input projection. This preserves spatial ordering while allowing permutation-invariant attention when needed.
\begin{algorithm}[tb]
\caption{Agent Initialization Training}
\label{alg:init_training}
\begin{algorithmic}[1]
\State \textbf{Input:} $\mathcal{M}, \Vec{\rvx}^0 \sim q(\Vec{\rvx}^0), \bar{\alpha_t}$
\For{each iteration $i = 1$ to $maxiter$}
    \State  $\gamma \sim \text{Bernoulli}(0.5)$, $t \sim \gU[0, \ldots, T]$, $\boldsymbol\epsilon \sim  \gN(\mathbf{0}, \rmI)$
    \State $\vec{\rvx}^{t} \leftarrow \sqrt{\bar{\alpha_t}}\vec{\rvx}^0 + \sqrt{1 - \bar{\alpha_t}}\boldsymbol\epsilon$
    
     \State $M_{a2a} \leftarrow \gamma \mathbf{1}_{n\times n} + (1 - \gamma) \mathbf{I}_n$
     \State $\gL_{\mathbb{I}}(\theta) \leftarrow ||\boldsymbol\epsilon - \boldsymbol\epsilon_{\theta}(\vec{\rvx}^t \odot  M_{a2a}, \mathcal{M}, t)||^2$
    \State $\theta \leftarrow \theta - \eta\nabla_\theta \gL_\mathbb{I}(\theta) $ 
\EndFor\\
\Return $\bm\epsilon_\theta$
\end{algorithmic}

\end{algorithm}

\begin{algorithm}[tb]
\caption{Agent Initialization Sampling}
\label{alg:init_sampling}
\begin{algorithmic}[1]
\State \textbf{Input: } $\mathcal{M}, N, \vec\rvx^{T}\sim  \gN(\mathbf{0}, \rmI)$
\For{each iteration $t \in \{T, T -1, \dotsc, 1\}$}
    \State $\hat{\boldsymbol\epsilon} \leftarrow  \boldsymbol\epsilon_{\theta}(\vec{\rvx}^t, \mathcal{M}, t)$
    \State $\vec\rvx^0 \leftarrow \frac{\vec\rvx^t - \sqrt{1 - \bar{\alpha}_t} \,\hat{\boldsymbol\epsilon}}{\sqrt{\bar{\alpha}_t}}
    $
    
    \State $\vec\rvx^{t-1} \leftarrow \sqrt{\bar{\alpha}_{t-1}} \, \vec\rvx^0 + \sqrt{1 - \bar{\alpha}_{t-1}} \, \hat{\boldsymbol\epsilon}
    $
\EndFor\\
\Return ${\vec\rvx^{0}}$
\end{algorithmic}
\end{algorithm}

\subsection{Trajectory Generation}
\label{sec:scene_traj}
Once agent initializations are obtained, we proceed to generate trajectories. As in Problem~\ref{prob:init}, we adopt a DDPM-based VLO objective, now applied over a low-dimensional latent representation of trajectories to reduce computational overhead. The denoising process is conditioned on the diffusion step $t$, map $\mathcal{M}$, agent initializations ${\Vec{\rvx}^0}$ (Section~\ref{sec:scene_init}), and agent type. To encourage diversity while ensuring realistic behavior, we additionally condition on a set of candidate trajectories $\mathcal{C}$ derived from the \gls{ff}, detailed in the next subsection. The corresponding training objective is given in~\Eqref{eq:traj_obj}.
\subsubsection{Latent Representation}  
While learning directly in trajectory space is possible, it results in (1) increased computational overhead and (2) high variance due to differences in map scale, position, and orientation. To mitigate this, we follow~\cite{jiang2023motiondiffusercontrollablemultiagentmotion} by applying Principal Component Analysis (PCA)~\cite{mackiewicz1993principal} for dimensionality reduction on local coordinate frame to enable efficient modeling of long-horizon motion. We denote the resulting latent representation as $\boldsymbol{\tau}_z$.
\subsubsection{Frenet Candidates}
While conditioning on map features alone can model the trajectory distribution, we observe that this often leads to unrealistic outputs e.g., straight-line trajectories that fail to follow curves or respect traffic semantics. To address this, we generate a set of candidate trajectories $\mathcal{C}$ using the Frenet Frame and treat them as motion primitives. 

\begin{figure}[htbp]
        \centering
        \includegraphics[width=\linewidth]{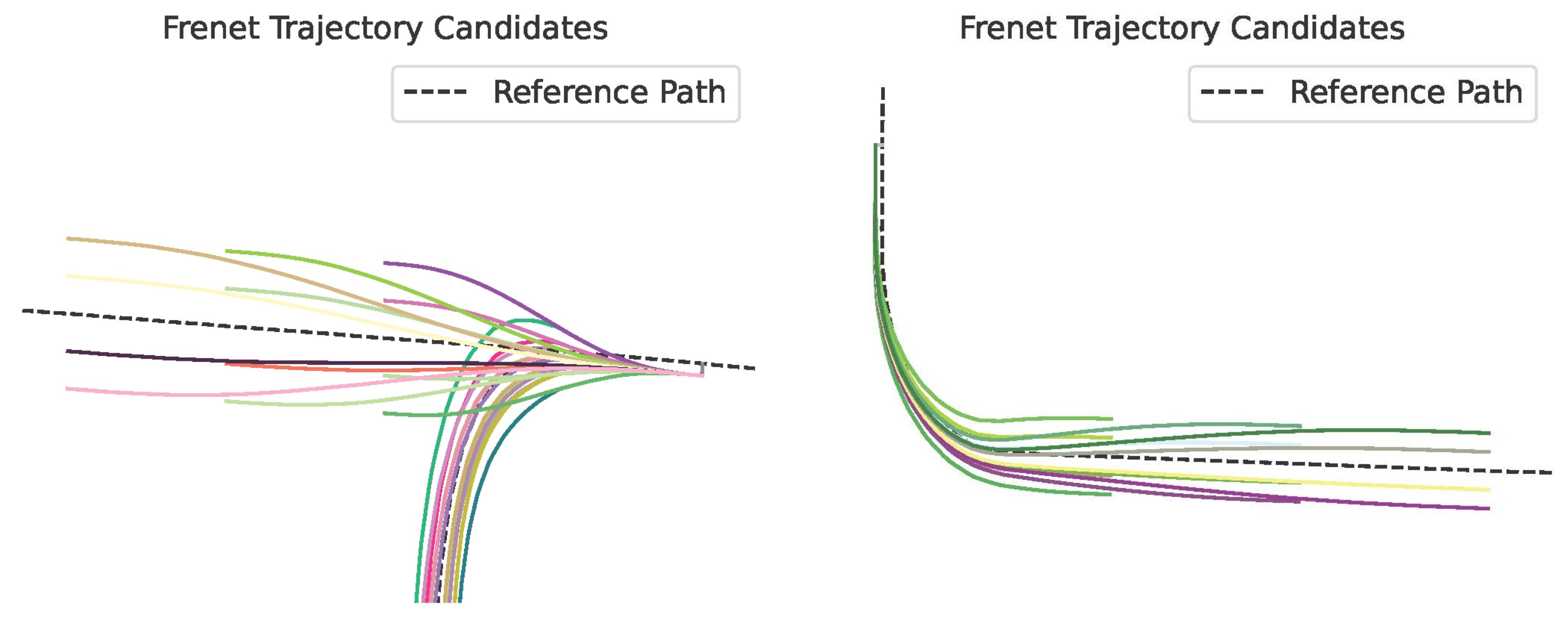}
    \caption{\textbf{Trajectory Candidates.} Candidates (colored) and reference lane (black). Multiple trajectory candidates around a reference path, are created by exploring a set of initial velocities and lateral deviations.}
    \label{fig:frenet}
\end{figure}

When an agent's initial position ${\Vec{\rvx}^0}$ is close to multiple lanes, all relevant lanes are included in $l_\mathcal{R}$ as illustrated in Fig.~\ref{fig:frenet}. As detailed in Algorithm~\ref{alg:frenet_cand}, diverse trajectory candidates are generated using the initial positions ${\Vec{\rvx}^0}$, reference lanes $l_\mathcal{R}$, and a predefined grid of initial speeds $\sV$ and lateral offsets $\sD$. These candidates improve trajectory diversity and reduce off-road violations by aligning the predictions with map topology. Similar to agent trajectories, we also transform the candidate trajectories into the latent space.
\begin{algorithm}[t]
\caption{Frenet Frame Candidate Generation}
\label{alg:frenet_cand}
\begin{algorithmic}[1]
\State \textbf{Input:} ($\mathcal{M}_\mathcal{R}, \hat{\vec\rvx}) \sim \text{Data}$, initial velocity set $\sV$, lateral deviation set $\sD$, time interval $\rvh$
\State $\gC \leftarrow \{\}$
\State Identify lanes $\rvl_{r}$ in which $\hat{\vec{\rvx}}$ lies
\For{each iteration $l_\mathcal{R} \in \rvl_\mathcal{R}$}
\State Compute $s_{0}$, $d_{0}$ based on $l_\mathcal{R}$
\For{each iteration $v_i \in \sV$, $d_H \in \sD$}
\State $(\rvs, \rvd) \leftarrow (s_{0} + v_i\rvh$, $d_{0} + d_H\rvh$)

\State {\color{gray}{\texttt{\#Frenet to Cartesian Coordinate}}}
\State $\boldsymbol{\tau}_{f} \leftarrow Fren2Cart(\rvs, \rvd)$
\State $\gC \leftarrow \gC \cup \{\boldsymbol{\tau}_{f} \}$
\EndFor
\EndFor\\
\Return $\gC $
\end{algorithmic}

\end{algorithm}

\subsubsection{\gls{hmp}}
The trajectory denoiser must capture both agent-agent and agent-map interactions. Each noisy latent trajectory $\boldsymbol{\tau}_z^t$ attends to its corresponding Frenet candidates $\mathcal{C}$, which are derived from the same initial state. To integrate multi-agent context, we alternate cross-attention and self-attention layers over agents and the vectorized map, enabling rich interaction modeling (see~\fig{fig:overall_diagram}C). The training objective for trajectory generation is then defined as:
\begin{equation}  \label{eq:traj_obj}
\min_{\varphi} \, \mathcal{L}_{\mathbb{T}}(\varphi) = 
\mathbb{E}\left[
\left\| \boldsymbol\epsilon - \bm\epsilon_\varphi\left(\boldsymbol{\tau}_z^t,\mathcal{M}, {\Vec{\rvx}^0}, \mathcal{C}_z, t \right) \right\|^2
\right]
\end{equation}

Trajectory training and sampling follow the same procedure as scene initialization (Algorithms~\ref{alg:init_training},~\ref{alg:init_sampling}).

\section{Experiments}
\label{sec:experiments}
Our experiments aim to answer the following questions:
\begin{itemize}
\item How does \gls{pd} compare with existing baselines in terms of distributional realism and common-sense metrics for agent initialization?
\item How does \gls{pd} perform compared to baselines on trajectory generation metrics?
\item How do the individual components of \gls{pd} influence overall performance?
\item How robust is the framework to map perturbations?
\end{itemize}
For additional visualizations and high-resolution plots, please refer to our project page.\footnotemark
\footnotetext{\href{https://github.com/CL2-UWaterloo/PathDiffuser/}{\textcolor{blue}{https://github.com/CL2-UWaterloo/PathDiffuser/}}}
\vspace{-2mm}

\subsection{Experimental Setup}
\begin{table*}[tb]
\centering
\caption{Agent Initialization Results}
\begin{tabular}{l|ccccc|ccc}
\hline
         & \multicolumn{5}{c|}{Distributional JSD $\textcolor{red}{\downarrow}$}                                           & \multicolumn{3}{c}{Common Sense Metrics $\textcolor{red}{\downarrow}$}                  \\ \cline{2-9} 
         & Near. Dist.    & Local Density  & Lat. Dev.      & Ang. Dev.     & Speed          & Collision Rate (\%) & Near. Edge (m) & Off-Road Rate (\%) \\ \hline
GT       &         -       &        -        &           -     &       -        &                & 0.70                 & 1.42          & 0.10                \\
MHA      & 0.29          & 0.14          & 0.20          & \textbf{0.11}          & 0.04          & 17.16               & \textbf{1.56}  & \textbf{4.26}      \\
PD $\ominus$ CDB & 0.15          & 0.08          & \textbf{0.20} & \textbf{0.11}          & 0.14          & 6.27                & 1.59           & 5.70                \\
PD (ours)      & \textbf{0.15} & \textbf{0.08} & \textbf{0.21} & 0.12 & \textbf{0.13} & \textbf{6.09}       & 1.59           & 5.77               \\ \hline
\end{tabular}

\label{tab:init_result}
\end{table*}

\begin{table*}[tb]
\centering
\caption{Trajectory Generation Results: Realism and Road Compliance}
\begin{tabular}{l|cccc|cccc}
\hline
       & \multicolumn{4}{c|}{No Map Perturbation}                                                  & \multicolumn{4}{c}{Map Perturbation}                                                     \\\cline{2-9} 
       & \multicolumn{2}{c|}{Realism $\textcolor{red}{\downarrow}$}                       & \multicolumn{2}{c|}{Road compliance $\textcolor{red}{\downarrow}$} & \multicolumn{2}{c|}{Realism $\textcolor{red}{\downarrow}$}                       & \multicolumn{2}{c}{Road compliance $\textcolor{red}{\downarrow}$} \\ \cline{2-9} 
       & actorCR       & \multicolumn{1}{c|}{Off-road Rate} & Avg. Lat. Dev.        & Final Lat. Dev.  & actorCR       & \multicolumn{1}{c|}{Off-road Rate} & Avg. Lat. Dev.       & Final Lat. Dev. \\ 
              & (\%)       & \multicolumn{1}{c|}{(\%)} & (m)        & (m)  & (\%)       & \multicolumn{1}{c|}{(\%)} & (m)         & (m) \\ \hline
GT     & 1.82          & \multicolumn{1}{c|}{5.49}          & 12.17             & 1.13             & 1.82          & \multicolumn{1}{c|}{5.49}          & 12.75           & 1.16           \\
VD     & 2.30           & \multicolumn{1}{c|}{\textbf{5.14}} & \textbf{11.77}            & 1.63            & \textbf{2.25} & \multicolumn{1}{c|}{\textbf{4.98}} & 12.24             & 2.48            \\
PD $\ominus$ P & \textbf{1.82} & \multicolumn{1}{c|}{5.67}          & 12.24           & \textbf{1.25}  & 2.40           & \multicolumn{1}{c|}{5.52}          & 13.01           & 1.90           \\
PD (ours)     & 2.01          & \multicolumn{1}{c|}{5.69}          & 11.78  & 1.37           & 2.40           & \multicolumn{1}{c|}{5.40}           & \textbf{12.14}  & \textbf{1.52} \\ \hline
\end{tabular}

\label{tab:traj_result}

\vspace{-5mm}
\end{table*}

\begin{table}[t]
\centering
\caption{Trajectory Generation Results: Groudtruth Comparison}
\begin{tabular}{l|ccc|ccc}
\hline
       & \multicolumn{3}{c|}{No Map Perturbation}       & \multicolumn{3}{c}{Map Perturbation}          \\  \cline{2-7} 
       & MR $\textcolor{red}{\downarrow}$             & FDE $\textcolor{red}{\downarrow}$     & ADE $\textcolor{red}{\downarrow}$     & MR $\textcolor{red}{\downarrow}$             & FDE $\textcolor{red}{\downarrow}$     & ADE $\textcolor{red}{\downarrow}$    \\
       
    &(\%) & (m) &(m) &(\%) &(m) & (m) \\ \hline

VD     & 52.21          & 21.61         & 7.84          & 53.44          & 23.33         & 8.32      
 \\
PD $\ominus$ P & \textbf{41.89} & \textbf{5.84} & \textbf{1.86} & \textbf{47.65} & \textbf{8.14} & \textbf{2.60} \\
 PD (ours)     & 43.84          & 7.16          & 2.57          & 48.69          & 8.63          & 3.07          \\ \hline
\end{tabular}

\label{tab:traj_result_gtcomparison}
\end{table}
\subsubsection{Dataset}
For both agent initialization and trajectory generation, we use the Argoverse 2 Motion Forecasting Dataset~\cite{wilson2023argoverse2generationdatasets}, a large-scale benchmark for evaluating motion prediction and planning in complex urban environments.

\subsubsection{Agent Initialization}
For agent initialization, the map is scaled so that the region which contains the agents is scaled to -1 to 1. Each agent's initial pose is represented by \(\vec\rvx\) and contains position \(x, y\), heading $\theta$ and initial speed. The position and initial speed attributes are normalized with minimum and maximum value from the training set.

\subsubsection{Trajectory Generation}

For evaluation, we consider a 6-second window of trajectories for each scene. To assess overall performance, we evaluate the agent initialization and trajectory generation models both independently and jointly, where the output of the initialization model is used to condition the trajectory generation model.

\subsection{Agent Initialization}

\subsubsection{Baselines}
We compare against SceneControl~\cite{lu2024scenecontrol}
, a diffusion-based framework for generating initial traffic scenes, which we refer to as \textbf{Multi-Head Attention~(\gls{mha})}. This baseline is trained using the objective defined in~\Eqref{eq:init_obj}. In contrast, our method \textbf{\gls{pd}} includes additional design components (Section~\ref{sec:scene_init}). We also evaluate an ablated variant, \textbf{\gls{pd}$\ominus$\gls{cdb}}, without \gls{dt} and \gls{cdb} to assess their impact. SceneControl outperforms popular autoregressive baselines such as those presented in~\cite{Tan_2021_CVPR}.
 
\subsubsection{Metrics}
To evaluate the realism of generated traffic behaviors, we report (1) \textit{Common-sense Metrics}: measured by the \textbf{Collision Rate}, which measures if the agents collided, distance to the nearest lane (\textbf{Near. Edge}), which measures the distance from the nearest center lane, and \textbf{Off-Road Rate}, which shows ratio of agents that are outside of the road. (2) \textit{Distributional Fidelity}: evaluated by using the Jensen-Shannon Divergence (\textbf{JSD}) between the generated and ground-truth distributions using histograms over key behavioral attributes as seen in~\cite{lu2024scenecontrol}. We report  inter-agent spacing via nearest-agent distance (\textbf{Near. Dist.}), and inter-space spacing of 5 closest agents (\textbf{Local Density}). Moreover, we assess alignment to road geometry via lateral deviation (\textbf{Lat. Dev.}) and angular deviation (\textbf{Ang. Dev.}). Lastly, we compare the overall speed distribution(\textbf{Speed}).

\begin{figure}[tp]
    \centering
    \includegraphics[width=\linewidth]{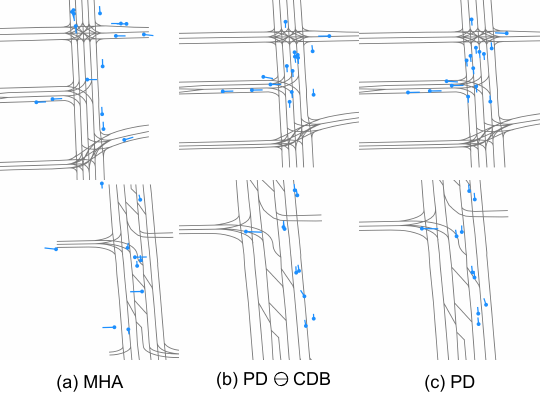}
    \caption{\textbf{Qualitative Results of Agent Initialization}. Two different map to compare the agent initializations. Each dot indicates the initial position and the lines indicates the heading.}
    \label{fig:init}
\vspace{-5mm}
\end{figure}

\subsubsection{Quantitative Results}
Table~\ref{tab:init_result} summarizes our results on the Argoverse 2 dataset. Overall, we observe that \textbf{\gls{pd}} outperforms the baselines across most metrics. Specifically, introducing \gls{dt} in \textbf{\gls{pd}$\ominus$\gls{cdb}} reduces the collision rate by approximately 10\%, albeit with a slight increase in off-road violations compared to \textbf{VD}, affecting map compliance. Further incorporating \gls{cdb}-based regularization improves distributional metrics. Our variants offer localized spatial attention without relying on costly gradient-based guided sampling.
\subsubsection{Qualitative Results}
\fig{fig:init} shows the quality of the generated initializations. \textbf{\gls{pd}} demonstrated the best map compliance and collisions compared to \gls{pd}$\ominus$\gls{cdb} and \gls{mha}. We can see that in \textbf{\gls{pd}$\ominus$\gls{cdb}}, map compliance improves while the collision still exists. Especially, initializations from \textbf{\gls{mha}} shows they are off-road and heading angle of some agents do not align with the road.

\subsection{Trajectory Generation}
Using the map and agent initializations, we generate trajectories that start from the given initial poses. 
\subsubsection{Baselines}
OptTrajDiff~\cite{wang2024optimizingdiffusionmodelsjoint} is most closely related to our work; however, it leverages agent history and frames the task as trajectory forecasting rather than generation. We refer to such baselines as \textbf{\gls{vd}}, which are trained using the objective in~\Eqref{eq:traj_obj}. In contrast, our approach \textbf{\gls{pd}} and its ablated variant \textbf{\gls{pd_p}} (which omits Frenet candidates) is designed for trajectory generation, as detailed in Section~\ref{sec:scene_traj}.

\begin{figure}[b]
    \centering
    \includegraphics[width=\linewidth]{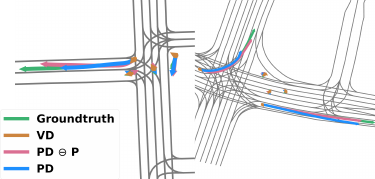}
    \caption{\textbf{Qualitative Results of Trajectory Generation}. Generated trajectories from \gls{vd}, \gls{pd_p}, and \gls{pd} from same map and overlayed for comparison.}
    \label{fig:traj}
\end{figure}

\begin{figure*}[ht]
    \centering
    \includegraphics[width=0.9\linewidth]{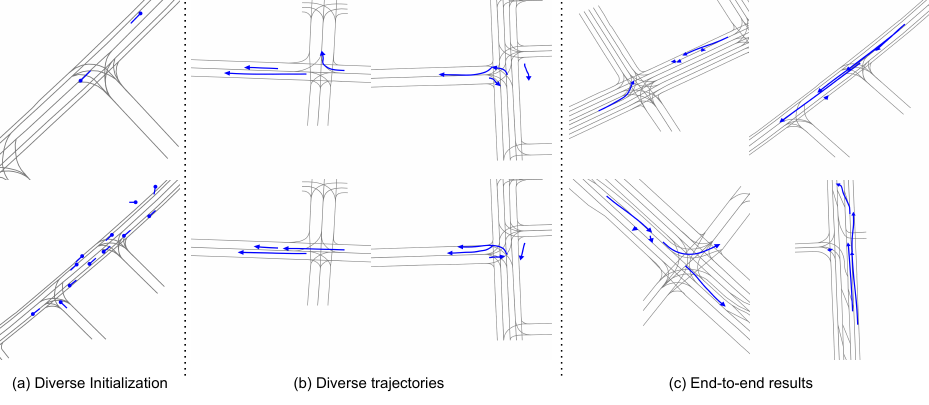}
    \caption{\textbf{Diverse Generation Result}. (a) Initialization visualization with different number of agents in the same map. (b) Trajectories of samples from the \gls{pd}. (c) End-to-end results which uses the generated initialization for trajectory generation.}
    \label{fig:diversity}
\vspace{-5mm}
\end{figure*}

\subsubsection{Metrics}
We evaluate the generated trajectories using the following criteria:
(1) \emph{Realism}: measured by the Actor Collision Rate (\textbf{actorCR}), which captures collisions between agents over the prediction horizon, and the \textbf{Off-road Rate}, which indicates whether trajectories deviate more than 4 meters from the road centerline.
(2) \emph{Road Compliance}: assessed by identifying reference lanes from each agent's initial position and computing the Average Lateral Deviation (\textbf{Avg. Lat. Dev.}) and Final Lateral Deviation (\textbf{Final Lat. Dev.}), which measure lateral deviation from the lane centerline across the prediction horizon $H$ and at the final timestep, respectively.
(3) \emph{Ground-Truth Comparison}: we report standard motion forecasting metrics, Final Displacement Error (\textbf{FDE}), which quantifies the average displacement error at the final timestep; Average Displacement Error (\textbf{ADE}), the mean displacement error over the entire horizon $H$; and Miss Rate (\textbf{MR}), defined as the percentage of predicted trajectories whose FDE exceeds 2 meters. Note that we focus on trajectory generation rather than reconstruction. Therefore:  
(1) Existing autoregressive models rely on agent history encoding, making direct comparison infeasible; hence, they are excluded from our evaluation.  
(2) For quantitative evaluation, we use ground-truth initialization and provide all models with scene-level context, including agent type and attributes such as initial speed, to enable realistic and context-aware trajectory generation.

\subsubsection{Quantitative Results}
Table~\ref{tab:traj_result} and Table~\ref{tab:traj_result_gtcomparison}, summarizes the results of trajectory generation. We can observe that \textbf{\gls{pd_p}} and \textbf{\gls{pd}} shows comparable result when there is no map perturbation. By adding an additional token to \gls{vd}, the model better captures agent dynamics compared to \gls{vd} and improved Final Lateral Deviation metric by approximately 0.4 meters. While \textbf{\gls{vd}} shows the best off-road rate, comparison to groundtruth shows that generated trajectories are not close from the groundtruth.

\subsubsection{Qualitative Results}
\fig{fig:traj} shows the how well generated trajectories are aligned to the map and how realistic each trajectories are. As in quantitative results, \textbf{\gls{pd}} and \textbf{\gls{pd_p}} generated trajectories that are well aligned to the map structure and does not collide with other agents. However, trajectories generated by \textbf{\gls{vd}} shows most of the trajectories are not moving from its initial position, which can explain why off-road rates are the lowest. 

\subsubsection{Diversity}
In~\fig{fig:diversity}, it highlights the model's capacity for producing diverse and varied traffic scenarios. With both agent initialization and trajectory generation models, we can generate novel traffic scenarios given a map which aligns with its structure. Specifically, we seek stochasticity; i.e., the ability to generate multiple plausible traffic scenarios for the same scene. From Tables~\ref{tab:traj_result} and~\ref{tab:traj_result_gtcomparison}, diversity can be quantified by observing higher ground-truth errors and lower map compliance, indicating that the model generates plausible alternatives that deviate from the reference.

\begin{figure}[b]
    \centering
    \includegraphics[width=\linewidth]{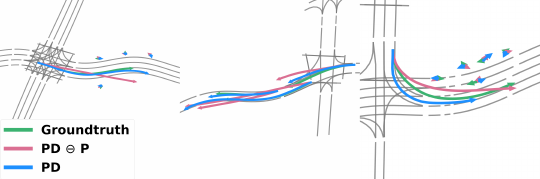}

\caption{\textbf{Scene-Attack Perturbations.} We leverage the Scene-Attack benchmark\cite{bahari2022sattack} to evaluate the robustness of our models under diverse, OOD map structures. The generation result with the map perturbation.}
    \label{fig:attack}
\end{figure}

\subsection{Robustness to Perturbations}

We evaluate the robustness of our model and baselines under map perturbations. As shown in~\fig{fig:attack}, the vectorized map is modified to test performance on out-of-distribution (OOD) map structures~\cite{bahari2022sattack}. 
Under OOD map scenarios, we observe that the \gls{pd_p} struggles due to the limited expressiveness of its low-dimensional PCA-based representations, often producing straight-line trajectories that fail to align with the perturbed map geometry.
However, the Frenet candidates remain robust to map perturbations by consistently aligning with the structure of the perturbed lane. We use a guided sampling process~\cite{dhariwal2021diffusion} to enforce alignment between predicted trajectories and Frenet-frame candidates. At each sampling step, we compute the distance to all candidates and select the closest one as the reference for alignment. Table~\ref{tab:traj_result} (map perturbations) shows that \gls{pd} outperforms baselines on road compliance metrics, largely due to the use of candidate trajectories generated in the Frenet frame. This conditioning anchors the generation process to the underlying map structure, which is also qualitatively evident in~\fig{fig:attack}. 

We observe that using Frenet-based candidates introduces meaningful stochasticity and enables the model to better capture road curvature, in contrast to baseline models that often default to straight-line paths.

\section{Conclusion}
\label{sec:conclusion}

We study the problem of generating realistic and diverse traffic scenarios that remain robust under map perturbations, ensuring both semantic consistency and physical plausibility for high-fidelity traffic simulation. We propose \model{}, a two-stage diffusion framework for agent initialization and trajectory generation. By leveraging \gls{dt} for canceling attention noise and introducing localized spatial attention, our model significantly reduces collision rates. Furthermore, removing history dependence and incorporating Frenet-frame-based candidates enables diverse trajectory generation beyond simple driving log replay. By decoupling agent initialization from trajectory generation, our method enhances both distributional fidelity and scene controllability. Extensive experiments show that our approach outperforms baselines in realism and diversity, while maintaining robustness to map perturbations.

\vspace{0.5em}
\noindent\textbf{Limitations and Future Work.} Due to the use of PCA-based dimensionality reduction, our current framework is limited to a fixed trajectory horizon of 6 seconds. In future work, we plan to explore more flexible representations that enable sampling of longer trajectories even when trained on shorter sequences. To assess error accumulation and traffic flow, we plan to evaluate the model in a closed-loop setting. 

Additionally, we plan to explore diffusion model variants that offer faster convergence and greater test-time controllability. Despite the use of dimensionality reduction, training remains computationally intensive, taking approximately 10 hours. To improve inference efficiency, we propose using Frenet candidates as a learned diffusion prior replacing the standard $\mathcal{N}(\mathbf{0}, \mathbf{I})$ initialization to reduce the number of diffusion steps required. Since the candidate grid is currently predefined, we also plan to investigate its effect in edge cases and its adaptability to complex scenarios. 

In the end-to-end process, trajectory generation model relies on the initialization output, making use of two models costly for both training and inference. As a future work, we plan to integrate the two models into an end-to-end framework to improve coherence and efficiency.

Finally, to improve the logical consistency of generated scenes, we intend to incorporate richer contextual cues such as traffic light signals, time of day, and weather conditions.




\bibliographystyle{ieee_fullname}
\bibliography{IEEEabrv,IEEEexample}


\end{document}